\documentclass[10pt]{article}
\usepackage{silence}
\usepackage[utf8]{inputenc}
\usepackage[T1]{fontenc}
\usepackage[charter]{mathdesign}
\usepackage[margin=1in]{geometry}
\usepackage{graphicx}
\usepackage{amsmath}
\usepackage{bm}
\usepackage{microtype}
\usepackage{array,booktabs}
\usepackage{float}
\usepackage{makecell}
\usepackage{xcolor}
\usepackage[cachedir=.]{minted}
\usepackage{parskip}
\usepackage{fancyhdr}
\definecolor{blue}{RGB}{54,125,189}
\usepackage[
    backend=biber,
    citestyle=numeric,
    bibstyle=numeric,
    mincitenames=1,
    maxcitenames=1,
    maxbibnames=3,
    doi=false,
    sorting=none,
]{biblatex}
\usepackage{xurl}
\usepackage[colorlinks=true,linkcolor=blue,citecolor=blue,urlcolor=blue]{hyperref}

\fancypagestyle{firstpage}{
  \fancyhf{}
  \fancyhead[L]{\small Preliminary Version; Work in Progress.}
  \fancyhead[R]{}
  \fancyfoot[C]{\thepage}
}

\begin{document}

\thispagestyle{firstpage}

\title{Mixture of Lookup Key-Value Experts}
\author{Zongcheng Wang \thanks{The School of Mathematics, Renmin University of China. Correspondence: 2021103675@ruc.edu.cn}}
\date{}

\maketitle
\thispagestyle{firstpage}

\fontsize{11pt}{13.2pt}\selectfont

\begin{abstract}
\fontsize{10.5pt}{12.6pt}\selectfont
Recent research has developed several LLM architectures suitable for inference on end-user devices, such as the Mixture of Lookup Experts (MoLE)~\parencite{jie_mixture_2025}. A key feature of MoLE is that each token id is associated with a dedicated group of experts. For a given input, only the experts corresponding to the input token id will be activated. Since the communication overhead of loading this small number of activated experts into RAM during inference is negligible, expert parameters can be offloaded to storage, making MoLE suitable for resource-constrained devices. However, MoLE's context-independent expert selection mechanism, based solely on input ids, may limit model performance. To address this, we propose the \textbf{\textcolor{black}{M}}ixture \textbf{\textcolor{black}{o}}f \textbf{\textcolor{black}{L}}ookup \textbf{\textcolor{black}{K}}ey-\textbf{\textcolor{black}{V}}alue Experts (\textbf{MoLKV}) model. In MoLKV, each expert is structured as a key-value pair. For a given input, the input-derived query interacts with the cached key-value experts from the current sequence, generating a context-aware expert output. This context-aware mechanism alleviates the limitation of MoLE, and experimental results demonstrate that MoLKV achieves significantly lower validation loss in small-scale evaluations.
\end{abstract}

\section{Introduction}

Currently, AI inference is mainly performed in cloud data centers. This approach benefits from high-performance hardware capable of running cutting-edge models. It also achieves cost efficiency through economies of scale and batch processing. An alternative is on-device AI, where models run locally on end-user devices like laptops and smartphones. Although hardware constraints limit end-user devices from running the most advanced models, continuous improvements in hardware capabilities and model capability density~\parencite{xiao_densing_2025} are expanding the scope of on-device AI. Furthermore, on-device AI has several unique advantages: 

\begin{enumerate}
    \item \textbf{Privacy Preservation}. By operating offline, it ensures that sensitive data remains on the device, thereby enhancing user privacy.
    \item \textbf{Personalization}. On-device AI models can learn from user data to achieve personalization while protecting privacy. Personalization with limited data can be achieved through in-context learning or fine-tuning, but long-term personalization requires advancements in continual learning and memory techniques. There has been some recent exploration in these areas~\parencite{eyuboglu_cartridges_2025, li_memos_2025}.
    \item \textbf{Proactive Intelligence}. On-device AI models can be deeply integrated with device systems, processing user activity information on the device in real time, and thus providing timely and proactive assistance~\parencite{lu_proactive_2025, yang_contextagent_2025}, overcoming the shortcomings of traditional LLM's reactive response.
\end{enumerate}

The biggest obstacle for end-user devices to run cutting-edge models is their limited Random Access Memory (RAM) capacity. Mixture-of-Experts (MoE)~\parencite{shazeer_outrageously_2017} are one of the most popular LLM architectures nowadays. For a given input, MoE activates only a subset of parameters, allowing the model to scale its parameter count while maintaining low computational cost. However, this parameter scaling leads to a massive memory footprint, with cutting-edge MoE models requiring between 200GB and 1TB of RAM in FP8 format~\parencite{yang_qwen3_2025, deepseek-ai_deepseek-v3_2025, team_kimi_2025}. This far exceeds the capacity of typical end-user devices, which usually have less than 32GB of RAM, making such models infeasible to run locally. In contrast, device storage is much larger, ranging from 256GB to over 1TB, allowing parameters to be offloaded to storage and selectively loaded during inference based on routing results~\parencite{xue_powerinfer-2_2024}. However, the number of parameters activated by each input in MoE is still relatively large, and the storage bandwidth is much lower than that of RAM. These factors make communication a bottleneck, thus greatly increasing latency.

A promising solution is to use architectures with lower activation ratios, such as the Mixture of Lookup Experts (MoLE)~\parencite{jie_mixture_2025}. In MoLE, each token id is associated with a dedicated group of experts. For a given input, only the experts corresponding to the input token id will be activated. The activation ratio can be as low as 1/100,000, far less than the 1/100 of MoE. Therefore, MoLE can minimize communication overhead, making it suitable for offloading parameters to storage to support on-device inference. However, MoLE underperforms comparably sized MoE models. This is partly because MoLE selects experts based solely on input ids, while MoE does so based on the routing results of the hidden state. The hidden state contains contextual information, while input ids do not. Therefore, MoLE's context-independent expert selection mechanism may limit model performance.

To address this, we propose the Mixture of Lookup Key-Value Experts (MoLKV) model, which builds on MoLE. In MoLKV, experts are structured as key-value pairs rather than single values. In addition to selecting experts based on input ids, MoLKV also obtains context-aware expert outputs. Specifically, for a given input, the input-derived query interacts with the cached key-value experts from the current sequence, resulting in a dynamically weighted expert output. MoLKV has the following two advantages:

\begin{enumerate}
    \item \textbf{Better model performance.} The context-aware expert output mechanism briefly introduced above alleviates the problem of context-independent expert selection in MoLE, thereby improving model performance.
    \item \textbf{Efficient batch inference.} Although MoLKV selects more experts for computation per input than MoLE, the extra experts are cached in RAM, thus avoiding the consumption of valuable storage bandwidth. Since batch inference is constrained by storage bandwidth, MoLKV still maintains efficient batch inference capabilities.
\end{enumerate}

We conducted experiments to validate that the performance of MoLKV outperforms that of MoLE. By controlling the activation parameters of both models at 197M and the total parameters at 1.65B, the results show that the validation loss of MoLKV is 0.03 lower than that of MoLE, verifying the effectiveness of MoLKV.

\section{Related Work}

\textbf{Sparse Mixture-of-Experts}\quad The widely used Mixture-of-Experts (MoE) architecture is more precisely termed Sparse Mixture-of-Experts (SMoE)~\parencite{shazeer_outrageously_2017}. For each input, SMoE activates only a subset of experts, maintaining low computational cost while scaling its parameters. Using fine-grained experts further improves performance~\parencite{dai_deepseekmoe_2024}, with corresponding scaling laws established~\parencite{ludziejewski_scaling_2024, yu_scaling_2025}. A key training challenge is expert load balancing, typically addressed through auxiliary loss functions~\parencite{fedus_switch_2022, qiu_demons_2025} or auxiliary-loss-free methods~\parencite{wang_auxiliary-loss-free_2024}.

\textbf{Memory Layers}\quad Memory Layers represent an extreme case of fine-grained SMoE, where each expert is a vector, the total number of experts can reach millions~\parencite{he_mixture_2024}, and the activation rate can be less than 0.1\%. For extremely large expert counts, product keys enable efficient routing~\parencite{lample_large_2019}. Unlike SMoEs, Memory Layers inherently achieve balanced expert utilization~\parencite{berges_memory_2025}. Although no standardized architecture has yet been established, recent studies have proposed diverse designs~\parencite{berges_memory_2025, huang_ultra-sparse_2025, huang_ultramemv2_2025}. The low activation ratios of Memory Layers make them theoretically suitable for on-device inference.

\textbf{Lookup-based Models}\quad Unlike SMoE and Memory Layers, which dynamically select experts via routing, Lookup-based models rely on direct lookups. Two main approaches have emerged: Mixture of Lookup Experts (MoLE)~\parencite{jie_mixture_2025}, which integrates per-layer experts retrieved by input ids, and methods that expand the input embedding layer using n-gram token embeddings~\parencite{yu_scaling_2025, huang_over-tokenized_2025}. Lookup-based models achieve sparser activations than Memory Layers, making them especially suitable for on-device deployment and even batch inference. Furthermore, the expert routing in Lookup-based models is fixed, thus eliminating the need for auxiliary losses to achieve load balancing, making them easier to train than SMoE.

\section{Model Architecture}

\subsection{Preliminary}

Since MoLKV is derived from MoLE~\parencite{jie_mixture_2025}, we will first introduce the architecture of MoLE. MoLE has different structures in training and inference modes. We will begin with the inference mode. Since the structure of each layer of MoLE is exactly the same, we will select one layer to explain in the following text and ignore the layer number.

\textbf{Inference Mode}\quad The experts of MoLE in each layer can be represented as $\{\{ \bm{v}_{i,n}\}_{n=1}^N\}_{i=1}^{|V|}$, where $|V|$ is the vocabulary size, $N$ is the number of experts for each id, and $\bm{v}_{i,n} \in \mathbb{R}^d$ is the $n$-th expert corresponding to id $i$, with $d$ being the hidden size. For a given input with id $i$, the experts selected via lookup operations are $\{ \bm{v}_{i,n}\}_{n=1}^N$. Therefore, for each input, $N$ experts are activated out of the total of $N|V|$ experts, resulting in a sparsity of $1/|V|$. With vocabulary size around 100,000, the activation ratio can be as low as 1/100,000, making MoLE suitable for on-device inference. Let $\bm{h} \in \mathbb{R}^d$ denote the FFN input and $i$ denote the input id, the FFN output $\bm{y}$ during inference is computed as:
\begin{align}
    \bm{y} & = \bm{h} + \operatorname{FFN}(\bm{h}) + \sum_{n=1}^{N} s_{n} \bm{v}_{i,n} \label{eq:infer} \\
    s_{n} & = \operatorname{Softmax}_n\left(\bm{h}^\top \bm{r}_n\right) = \frac{\exp(\bm{h}^\top \bm{r}_n)}{\sum_{j=1}^{N} \exp(\bm{h}^\top \bm{r}_j)} \label{eq:routing}
\end{align}
According to \eqref{eq:infer}, the selected experts $\{\bm{v}_{i,n}\}_{n=1}^N$ are aggregated according to the routing scores $\bigl\{s_n \bigr\}_{n=1}^N$, and the resulting expert output is added to the original FFN output to produce the final output $\bm{y}$. The routing scores are computed by \eqref{eq:routing}, where $\bm{r}_{n} \in \mathbb{R}^d$ denotes the router vector for the $n$-th expert and is independent of input id.

\textbf{Training Mode}\quad The modification to the training mode involves replacing $\bm{v}_{i,n}$ with $\operatorname{FFN}_n(\bm{e}_i)$, where $\bm{e}_i \in \mathbb{R}^d$ denotes the token embedding for id $i$, and $\operatorname{FFN}_n(\cdot)$ represents the $n$-th FFN expert, which is independent of the input id. The computation is as follows:
\begin{align}
    \bm{y} & = \bm{h} + \operatorname{FFN}(\bm{h}) + \sum_{n=1}^{N} \left(s_{n} \operatorname{FFN}_n(\bm{e}_i)\right) \label{eq:train}
\end{align}
In training mode, each layer of MoLE has $N$ FFN experts, denoted as $\{\operatorname{FFN}_n(\cdot)\}_{n=1}^N$. For a given input with id $i$, the corresponding token embedding $\bm{e}_i$ is used as the input to the FFN experts, and the resulting outputs for id $i$ are $\{\operatorname{FFN}_n(\bm{e}_i)\}_{n=1}^N$, as shown in \eqref{eq:train}. After training, the weights of $\bm{e}_i$ and $\operatorname{FFN}_n(\cdot)$ are fixed, making $\operatorname{FFN}_n(\bm{e}_i)$ a static vector. This means we can pre-compute and store the results, denoted as $\bm{v}_{i,n}:=\operatorname{FFN}_n(\bm{e}_i)$. The above reparameterization transforms MoLE into inference mode.

\textbf{Analysis of the Two Modes}\quad Why does MoLE use different structures for training and inference? Suppose we use the inference-mode structure for training, as shown in \eqref{eq:infer}. The experts $\{ \bm{v}_{i,n}\}_{n=1}^N$ for id $i$ do not contain contextual information, only representing certain meanings of the token corresponding to id $i$. This implies that each $\bm{v}_{i,n}$ has a strong correlation with the corresponding token embedding $\bm{e}_i$ during training. Switching to training mode, which means representing $\bm{v}_{i,n}$ as $\operatorname{FFN}_n(\bm{e}_i)$, can introduce beneficial inductive biases without losing expressive power, thereby improving training dynamics. Our ablation studies further confirmed that replacing $\bm{v}_{i,n}$ with $\operatorname{FFN}_n(\bm{e}_i)$ during training leads to lower validation loss. However, since all FFN experts are activated in training mode, the number of activated parameters is very large, which leads to inefficient inference and makes parameter offloading impossible. Therefore, MoLE needs to be switched to inference mode after training to achieve efficient inference.

\textbf{Gated MoLE}\quad Inspired by Gated Attention~\parencite{qiu_gated_2025}, we enhance MoLE by adding a gating mechanism to the expert outputs, resulting in a stronger baseline termed Gated MoLE. The inference-mode computation for Gated MoLE is:
\begin{align}
    \bm{y} & = \bm{h} + {\operatorname{FFN}\left( \bm{h} \right)} + g\sum_{n=1}^{N} {s_{n} \bm{v}_{i,n}} \label{eq:gated mole} \\
    g &= \operatorname{Sigmoid}\left(\bm{h}^{\top}\bm{u}\right) \label{eq:gate}
\end{align}
The gating score $g$ is multiplied by the expert output, as shown in \eqref{eq:gated mole}. $g$ is computed by \eqref{eq:gate}, where $\bm{u} \in \mathbb{R}^d$ is the gate parameter. Gated MoLE introduces only $d$ additional parameters per layer, minimally increasing parameter count and computational cost, while improving performance through dynamic gating of the expert contributions.

\subsection{MoLKV}
The following text introduces our proposed MoLKV model. We will start from the motivation, explore step by step, and finally derive the architecture of MoLKV.

\textbf{Key-Value Experts}\quad The previous section detailed the architecture of MoLE. One limitation of MoLE is that its expert selection relies solely on input ids. Although the routing scores shown in \eqref{eq:routing} yield dynamically weighted expert outputs, the context awareness is limited because only a small number of experts are selected per input. A potential improvement is to expand expert selection beyond input token lookups, thereby engaging more experts. A natural approach is to associate each expert with a key, allowing experts to be selected based on similarity computations between these keys and input-derived queries. Therefore, MoLKV extends MoLE by associating each expert $\bm{v}_{i,n} \in \mathbb{R}^d$ with a key $\bm{k}_{i,n} \in \mathbb{R}^{d'}$ (where $d'$ denotes the key dimension), forming a key–value paired expert, or KV expert for short. Here, $\bm{k}_{i,n}$ and $\bm{v}_{i,n}$ are referred to as the expert key and expert value, respectively.

\textbf{Subset-of-Experts Problem}\quad After adding the expert keys, we can calculate the similarity scores between the query and the whole expert keys $\{\{ \bm{k}_{i,n}\}_{n=1}^N\}_{i=1}^{|V|}$, and select the expert values corresponding to the top-$k$ highest scores, which is similar to the way experts are selected in MoE. However, the number of expert keys per layer is as high as $N|V|$, where $|V|$ is around 100,000. Since the cost of computing with all expert keys is prohibitively high, we can only perform computations on a small subset of experts. This raises a problem: how can we effectively identify a suitable subset-of-experts for computation? If we randomly select some experts as the subset, it will obviously lead to suboptimal performance. A more reasonable approach is to use approximate nearest neighbor (ANN) search algorithms, such as HNSW~\parencite{malkov_efficient_2020}, which can retrieve expert keys that are highly similar to the query at a lower cost. However, this would introduce considerable training complexity and is not explored in this work. Moreover, under MoLE's inference mode, all experts are offloaded to storage. The methods described above would load much more experts from storage, consuming valuable storage bandwidth and potentially increasing significant latency, especially in batch inference scenarios. Therefore, these approaches represent suboptimal solutions.

\textbf{Subset-of-Experts Solution: Cached KV Experts}\quad This paper employs an alternative method: instead of choosing subset-of-experts from storage, we use the experts cached in RAM from the current sequence as the subset-of-experts. Specifically, during the inference phase, each token activates the experts corresponding to its input id, loading them from storage into RAM. We can cache the KV experts corresponding to the tokens in the current sequence, treating them as the subset-of-experts. In this way, for a new input in the sequence, it can interact with the cached KV experts to obtain new expert outputs, instead of interacting with all experts. It is worth mentioning that the cached KV experts are quite similar to the KV cache in attention, but the former are trainable parameters, while the latter consists of activations. The above method has two advantages:
\begin{enumerate}
    \item The cached KV experts are in RAM and do not need to be loaded from storage, thus avoiding the consumption of storage bandwidth. Therefore, it does not increase latency significantly and enables efficient batch inference.
    \item Similar to the KV cache, the cached KV experts contain contextual information, enabling the generation of effective, context-aware expert outputs.
\end{enumerate}
\textbf{Cached KV Experts in Detail}\quad The previous text introduced the concept of cached KV experts. Here we will detail its mathematical form. Similar to the sliding window attention, we only retain the experts corresponding to the most recent $M$ tokens preceding the current token (assuming the sequence is sufficiently long), resulting in preliminary cached KV experts $\bm{K} \in \mathbb{R}^{M \times N \times d'}$ and $\bm{V} \in \mathbb{R}^{M \times N \times d}$. Similar to positional encoding in attention, we apply RoPE~\parencite{su_roformer_2024} to $\bm{K}$ along the sequence dimension to obtain $\bm{K}^R$. An optional step is to normalize the cached KV experts; additionally, we flatten $\bm{V}$ along the first two dimensions to obtain $\bm{V}'$. We obtain the final cached KV experts as $\bm{K}^R \in \mathbb{R}^{M \times N \times d'}$ and $\bm{V}' \in \mathbb{R}^{MN \times d}$.

\textbf{MoLKV Architecture}\quad In the MoLKV model, let $\bm{h} \in \mathbb{R}^d$ denote the FFN input, and $i$ denote the input id corresponding to $\bm{h}$. The FFN output $\bm{y}$ during inference is computed as:
\begin{align}
    \bm{y} & = \bm{h} + \operatorname{FFN}(\bm{h}) + g\sum_{n=1}^{N} s_{n} \bm{v}_{i,n} + g'\bm{S}'_I\bm{V}'_I \label{eq:molkv}
\end{align}
where $g'\bm{S}'_I\bm{V}'_I$ is the new gated expert output introduced by MoLKV compared to Gated MoLE formulation in \eqref{eq:gated mole}. The new gating score $g'$ is calculated using $g' = \operatorname{Sigmoid}(\bm{h}^{\top}\bm{u}')$, with $\bm{u}' \in \mathbb{R}^d$ as the new gate parameter. The steps for obtaining the new expert output $\bm{S}'_I\bm{V}'_I$ are detailed below:

$\vartriangleright$ 1. The query $\bm{q} \in \mathbb{R}^{d'}$ is derived from $\bm{h}$ via a projection matrix $\bm{W}_q \in \mathbb{R}^{d'\times d}$. Then we can obtain the RoPE-encoded query $\bm{q}^R$:
\begin{align}
    \bm{q} & = \bm{W}_q \bm{h},\quad \bm{q}^R = \operatorname{RoPE}\left(\bm{q}\right)
\end{align}

$\vartriangleright$ 2. The new scores $\bm{S}\in \mathbb{R}^{MN}$ are obtained by summing the query-key scores $\bm{S}_{\text{qk}}$ and the new routing scores $\bm{S}_{\text{router}}$, then flattening:
\begin{align}
    \bm{S} = \operatorname{Flatten}\left(\bm{S}_{\text{qk}} + \bm{S}_{\text{router}}\right), \quad \bm{S}_{\text{qk}} & = \frac{\bm{K}^R\bm{q}^R} {\sqrt{d'}}, \quad \bm{S}_{\text{router}} = \bm{W}_r\bm{h}
\end{align}
similar to the multi-head attention scores, the qk scores $\bm{S}_{\text{qk}}\in \mathbb{R}^{M\times N}$ are obtained by multiplying the cached expert keys $\bm{K}^R$ by the RoPE-encoded query $\bm{q}^R$ and scaling by $1/\sqrt{d'}$. The route scores $\bm{S}_{\text{router}}\in \mathbb{R}^{1\times N}$ are obtained via a new router $\bm{W}_r \in \mathbb{R}^{1\times N\times d}$.

$\vartriangleright$ 3. Following sparse attention methods~\parencite{lu_moba_2025, deepseekai2025deepseekv32}, we apply a top-$k$ selection to reduce computation:
\begin{align}
    I & = \operatorname{SelectTopkIndices}\left(\bm{S}, k\right)\ \rightarrow\ \bm{S}_I \in \mathbb{R}^{k}, \ \bm{V}'_I \in \mathbb{R}^{k \times d}
\end{align}
where $I$ denote the indices of the top-$k$ values in $\bm{S}$. Using indices $I$, we obtain the top-$k$ scores $\bm{S}_I$ and corresponding cached expert keys $\bm{V}'_I$. After obtaining the normalized top-$k$ scores $\bm{S}'_I \in \mathbb{R}^{k}$ via $\bm{S}_I'=\operatorname{Softmax}\left(\bm{S}_I\right)$, the new gated expert output is computed as $g'\bm{S}_I'\bm{V}'_I \in \mathbb{R}^{k}$, as shown in \eqref{eq:molkv}. The attention-like interaction between the query and the cached KV experts enables the resulting expert output to be context-aware.

Additionally, the original routing score $s_{n}$ in \eqref{eq:molkv} is by augmenting the result from \eqref{eq:routing} with the score obtained from the query and the current expert key $\bm{k}_{i, n}$:
\begin{align}
    s_{n} & = \operatorname{Softmax}_n\left(\bm{h}^\top \bm{r}_n + \frac{\bm{q}^\top \bm{k}_{i, n}}{\sqrt{d'}}\right), \ s_{n} \in \mathbb{R}
\end{align}

\subsection{Complexity Analysis}

\begin{table}[h]
    \setlength{\tabcolsep}{7.5pt}
    \centering
    \caption{\small Complexities of Different Models. We report the statistics of a single FFN layer during the inference phase, and we only count the MACs of large matrix operations in these models.}
    \vskip 0.1in
    \scalebox{0.875}{
        \begin{tabular}{l|cccc}
            \toprule
            Models & MACs & \# Param in RAM & \# Param Offloaded & \# Param Loaded per Input \\
            \cmidrule(lr){2-2} \cmidrule(lr){3-3} \cmidrule(lr){4-4} \cmidrule(lr){5-5}
            & \textcolor{gray}{Computational Cost} & \textcolor{gray}{RAM Usage} & \textcolor{gray}{Storage Usage} & \textcolor{gray}{Storage Bandwidth Usage} \\
            \midrule
            Dense & $3dD$ & $3dD$ & 0 & 0 \\
            MoLE & $3dD$ & $3dD$ & $N|V|d$ & $Nd$ \\
            MoLKV & $3dD+dd'+MNd'+kd$ & $3dD+MN(d+d')$ & $N|V|(d+d')$ & $N(d+d')$ \\
            \bottomrule
        \end{tabular}
    }
    \label{tab:complexity}
\end{table}

During the inference phase, the complexity of a single FFN layer in the Dense, MoLE, and MoLKV models is summarized in Table~\ref{tab:complexity}. We report four important metrics. The meaning of each metric is explained in gray below its name. Here, $d$ is the hidden size, $D$ is the intermediate size of FFN, $d'$ is the size of the expert key, $N$ is the number of experts for each id, $|V|$ is the vocabulary size, and $M$ is the length of the cached experts. We used the SwishGLU form~\parencite{shazeer_glu_2020} of FFN, so the number of parameters for each FFN is $3dD$.

For the Dense model, which is the standard Transformer model, all its parameters are in RAM, so there is no storage usage or loading overhead. For MoLE, all experts are offloaded to storage, the total offloaded parameter count is $N|V|d$, and the experts loaded per input has a size of $Nd$.

For MoLKV, since each expert is a key-value pair, its size is $d+d'$, making the total offloaded parameter count $N|V|(d+d')$. Since it selects additional experts in RAM rather than storage, the number of experts loaded per input remains $N$, and the parameter count is $N(d+d')$. MoLKV needs to cache experts of length $M$ in RAM, so the number of parameters in RAM increases by $MN(d+d')$. The  computational cost of MoKLV represented in MACs is $3dD+dd'+MNd'+kd$, where $dd'$ corresponds to the query projection, $MNd'$ corresponds to the similarity calculation between the query and the expert keys, and $kd$ corresponds to the output projection of top-$k$ expert values. Due to the addition of the expert keys, MoLKV has a larger overhead compared to MoLE, but this issue can be mitigated by selecting a smaller $d'$ and $M$.

\section{Experiments}

\begin{table}[h]
    \centering
    \begin{minipage}{0.56\textwidth}
        \centering
        \caption{\small\text{Model Configurations}}
        \vskip 0.1in
        \scalebox{0.9}{
            \begin{tabular}{l|ccccccc}
                \toprule
                Models & $L$ & $d$ & $D$ & $N$  & \makecell[c]{\# Activated\\ Param} & \makecell[c]{\# Total\\ Param} \\
                \midrule
                Dense & 16 & 1024 & 2644 & 0 & 197M & 197M \\
                MoLE & 16 & 1024 & 2644 & 2 & 197M & 1.65B \\
                MoLKV & 16 & 1024 & 2548 & 2 & 197M & 1.65B \\
                \bottomrule
            \end{tabular}
        }
        \label{tab:model configs}
    \end{minipage}
    \hspace{-0.5cm}
    \begin{minipage}{0.42\textwidth}
        \centering
        \caption{\small\text{Model Performances}}
        \vskip 0.1in
        \scalebox{0.9}{
            \begin{tabular}{l|c}
                \toprule
                Models & Validation Loss \\
                \midrule
                Dense & 3.1083 \\
                MoLE & 3.0297 \\
                Gated MoLE & 3.0180 \\
                MoLKV & \textbf{2.9985} \\
                \bottomrule
            \end{tabular}
        }
        \label{tab:model performances}
    \end{minipage}
\end{table}

\textbf{Model Configurations}\quad We implement Dense, MoLE, and MoLKV models for experiments, and their configurations are shown in Table~\ref{tab:model configs}. The Dense model adopts the modern Transformer~\parencite{vaswani_attention_2017} architecture, including RoPE~\parencite{su_roformer_2024}, SwishGLU~\parencite{shazeer_glu_2020}, RMSNorm~\parencite{zhang_root_2019}, and Pre-LN~\parencite{xiong_layer_2020}. MoLE and MoLKV are built upon the Dense model by adding new modules. All three models consist of 16 layers with a hidden size of 1024. Both MoLE and MoLKV have 2 experts per token id at each layer. For MoLKV, the expert key size $d'$ is 146, the cached expert length $M$ is 512, the $k$ (used for selecting top-$k$ expert values) is set to 32, and only the first 14 layers contain experts. The FFN intermediate size $D$ of MoLKV is set to 2548, which is smaller than the 2644 of the other two. Under the above configuration, the three models share the same activation parameter count of 197M, and MoLE and MoLKV have the same total parameter count of 1.65B. This ensures a fair comparison in the experiments.

\textbf{Training Configurations}\quad All models are trained on a 10B-token subset of FineWeb dataset~\parencite{penedo_fineweb_2024} using the tiktoken tokenizer with a vocabulary size of 50304. We use AdamW~\parencite{loshchilov_adamw_2019} as the optimizer for all models, with a learning rate of 3e-4. The sequence length is 2048, the batch size is 240, and the training consists of 20,000 steps. Additional training hyperparameters are provided in Table~\ref{tab:training config}.

\textbf{Main Results}\quad As shown in Table~\ref{tab:model performances}, we report validation loss as the metric for evaluating model performance. Due to the introduction of the lookup experts, the validation loss of MoLE decreased by nearly 0.08 compared to that of Dense. MoLKV further reduced the loss by 0.03 relative to MoLE. To eliminate the effect of the gating mechanism in MoLKV, we also included Gated MoLE for comparison. MoLKV achieved a 0.02 lower validation loss than Gated MoLE, confirming the effectiveness of its core design. The experimental results demonstrate that the context-aware expert output mechanism introduced in MoLKV can effectively improve model performance.

\textbf{Upcoming Work}\quad Due to time and resource constraints, we have only conducted preliminary experiments as described. In the future, we will also conduct the following experiments: (1) Training on a larger dataset to enhance the persuasiveness of the experiments; (2) Evaluating on more benchmarks; (3) Ablation studies on the effectiveness of specific elements in MoLKV; (4) Testing the efficiency of MoLKV in inference, such as single-step decoding latency and throughput; (5) Conducting in-depth experimental analysis of MoLKV. After completing the above experiments, we will update this paper.

\section{Conclusion} This paper addresses the potential limitations of MoLE’s context-independent expert selection and proposes an improved architecture named MoLKV. Building upon MoLE, MoLKV enhances model performance by enabling queries to interact with cached key-value experts, thereby generating context-aware expert outputs while retaining MoLE’s advantage of efficient batch inference. This work advances the design of architectures suitable for on-device inference, contributing to the development of edge AI.

\newpage
\printbibliography

@inproceedings{vaswani_attention_2017,
	title = {Attention is {All} you {Need}},
	volume = {30},
	url = {https://proceedings.neurips.cc/paper/2017/hash/3f5ee243547dee91fbd053c1c4a845aa-Abstract.html},
	booktitle = {Advances in {Neural} {Information} {Processing} {Systems}},
	author = {Vaswani, Ashish and Shazeer, Noam and Parmar, Niki and Uszkoreit, Jakob and Jones, Llion and Gomez, Aidan N and Kaiser, Ł ukasz and Polosukhin, Illia},
	year = {2017},
}

@article{fedus_switch_2022,
	title = {Switch {Transformers}: {Scaling} to {Trillion} {Parameter} {Models} with {Simple} and {Efficient} {Sparsity}},
	volume = {23},
	issn = {1533-7928},
	url = {http://jmlr.org/papers/v23/21-0998.html},
	number = {120},
	journal = {Journal of Machine Learning Research},
	author = {Fedus, William and Zoph, Barret and Shazeer, Noam},
	year = {2022},
	pages = {1--39},
}

@misc{wang_auxiliary-loss-free_2024,
	title = {Auxiliary-{Loss}-{Free} {Load} {Balancing} {Strategy} for {Mixture}-of-{Experts}},
	url = {http://arxiv.org/abs/2408.15664},
	author = {Wang, Lean and Gao, Huazuo and Zhao, Chenggang and Sun, Xu and Dai, Damai},
	year = {2024},
}

@misc{lu_moba_2025,
	title = {{MoBA}: {Mixture} of {Block} {Attention} for {Long}-{Context} {LLMs}},
	url = {http://arxiv.org/abs/2502.13189},
	author = {Lu, Enzhe and Jiang, Zhejun and Liu, Jingyuan and Du, Yulun and Jiang, Tao and Hong, Chao and Liu, Shaowei and He, Weiran and Yuan, Enming and Wang, Yuzhi and Huang, Zhiqi and Yuan, Huan and Xu, Suting and Xu, Xinran and Lai, Guokun and Chen, Yanru and Zheng, Huabin and Yan, Junjie and Su, Jianlin and Wu, Yuxin and Zhang, Neo Y. and Yang, Zhilin and Zhou, Xinyu and Zhang, Mingxing and Qiu, Jiezhong},
	year = {2025},
}

@article{su_roformer_2024,
	title = {{RoFormer}: {Enhanced} transformer with {Rotary} {Position} {Embedding}},
	volume = {568},
	issn = {0925-2312},
	url = {https://www.sciencedirect.com/science/article/pii/S0925231223011864},
	journal = {Neurocomputing},
	author = {Su, Jianlin and Ahmed, Murtadha and Lu, Yu and Pan, Shengfeng and Bo, Wen and Liu, Yunfeng},
	year = {2024},
	pages = {127063},
}

@article{lu_proactive_2025,
	title = {Proactive {Agent}: {Shifting} {LLM} {Agents} from {Reactive} {Responses} to {Active} {Assistance}},
	volume = {2025},
	url = {https://proceedings.iclr.cc/paper_files/paper/2025/hash/75c37811e830bf029584b1c6fac17726-Abstract-Conference.html},
	journal = {International Conference on Representation Learning},
	author = {Lu, Yaxi and Yang, Shenzhi and Qian, Cheng and Chen, Guirong and Luo, Qinyu and Wu, Yesai and Wang, Huadong and Cong, Xin and Zhang, Zhong and Lin, Yankai and Liu, Weiwen and Wang, Yasheng and Liu, Zhiyuan and Liu, Fangming and Sun, Maosong},
	year = {2025},
	pages = {47431--47457},
}

@misc{li_memos_2025,
	title = {{MemOS}: {An} {Operating} {System} for {Memory}-{Augmented} {Generation} ({MAG}) in {Large} {Language} {Models}},
	url = {http://arxiv.org/abs/2505.22101},
	author = {Li, Zhiyu and Song, Shichao and Wang, Hanyu and Niu, Simin and Chen, Ding and Yang, Jiawei and Xi, Chenyang and Lai, Huayi and Zhao, Jihao and Wang, Yezhaohui and Ren, Junpeng and Lin, Zehao and Huo, Jiahao and Chen, Tianyi and Chen, Kai and Li, Kehang and Yin, Zhiqiang and Yu, Qingchen and Tang, Bo and Yang, Hongkang and Xu, Zhi-Qin John and Xiong, Feiyu},
	year = {2025},
}

@misc{eyuboglu_cartridges_2025,
	title = {Cartridges: {Lightweight} and general-purpose long context representations via self-study},
	url = {http://arxiv.org/abs/2506.06266},
	author = {Eyuboglu, Sabri and Ehrlich, Ryan and Arora, Simran and Guha, Neel and Zinsley, Dylan and Liu, Emily and Tennien, Will and Rudra, Atri and Zou, James and Mirhoseini, Azalia and Re, Christopher},
	year = {2025},
}

@article{xiao_densing_2025,
	title = {Densing law of {LLMs}},
	volume = {7},
	issn = {2522-5839},
	url = {https://www.nature.com/articles/s42256-025-01137-0},
	number = {11},
	journal = {Nature Machine Intelligence},
	author = {Xiao, Chaojun and Cai, Jie and Zhao, Weilin and Lin, Biyuan and Zeng, Guoyang and Zhou, Jie and Zheng, Zhi and Han, Xu and Liu, Zhiyuan and Sun, Maosong},
	year = {2025},
	pages = {1823--1833},
}

@misc{yang_contextagent_2025,
	title = {{ContextAgent}: {Context}-{Aware} {Proactive} {LLM} {Agents} with {Open}-{World} {Sensory} {Perceptions}},
	url = {http://arxiv.org/abs/2505.14668},
	author = {Yang, Bufang and Xu, Lilin and Zeng, Liekang and Liu, Kaiwei and Jiang, Siyang and Lu, Wenrui and Chen, Hongkai and Jiang, Xiaofan and Xing, Guoliang and Yan, Zhenyu},
	month = oct,
	year = {2025},
}

@article{malkov_efficient_2020,
	title = {Efficient and {Robust} {Approximate} {Nearest} {Neighbor} {Search} {Using} {Hierarchical} {Navigable} {Small} {World} {Graphs}},
	volume = {42},
	issn = {1939-3539},
	url = {https://ieeexplore.ieee.org/abstract/document/8594636},
	number = {4},
	journal = {IEEE Transactions on Pattern Analysis and Machine Intelligence},
	author = {Malkov, Yu A. and Yashunin, D. A.},
	year = {2020},
	pages = {824--836},
}

@misc{he_mixture_2024,
	title = {Mixture of {A} {Million} {Experts}},
	url = {http://arxiv.org/abs/2407.04153},
	author = {He, Xu Owen},
	year = {2024},
}

@article{penedo_fineweb_2024,
	title = {The {FineWeb} {Datasets}: {Decanting} the {Web} for the {Finest} {Text} {Data} at {Scale}},
	volume = {37},
	url = {https://proceedings.neurips.cc/paper_files/paper/2024/hash/370df50ccfdf8bde18f8f9c2d9151bda-Abstract-Datasets_and_Benchmarks_Track.html},
	journal = {Advances in Neural Information Processing Systems},
	author = {Penedo, Guilherme and Kydlíček, Hynek and Allal, Loubna B. and Lozhkov, Anton and Mitchell, Margaret and Raffel, Colin and Von Werra, Leandro and Wolf, Thomas},
	year = {2024},
	pages = {30811--30849},
}

@inproceedings{jie_mixture_2025,
	title = {Mixture of {Lookup} {Experts}},
	url = {https://proceedings.mlr.press/v267/jie25b.html},
	booktitle = {Proceedings of the 42nd {International} {Conference} on {Machine} {Learning}},
	author = {Jie, Shibo and Tang, Yehui and Han, Kai and Li, Yitong and Tang, Duyu and Deng, Zhi-Hong and Wang, Yunhe},
	year = {2025},
	note = {ISSN: 2640-3498},
	pages = {27929--27940},
}

@misc{deepseekai2025deepseekv32,
      title={DeepSeek-V3.2: Pushing the Frontier of Open Large Language Models}, 
	  url = {https://huggingface.co/deepseek-ai/DeepSeek-V3.2/blob/main/assets/paper.pdf},
      author={DeepSeek-AI},
      year={2025},
}

@misc{qiu_gated_2025,
	title = {Gated {Attention} for {Large} {Language} {Models}: {Non}-linearity, {Sparsity}, and {Attention}-{Sink}-{Free}},
	url = {http://arxiv.org/abs/2505.06708},
	author = {Qiu, Zihan and Wang, Zekun and Zheng, Bo and Huang, Zeyu and Wen, Kaiyue and Yang, Songlin and Men, Rui and Yu, Le and Huang, Fei and Huang, Suozhi and Liu, Dayiheng and Zhou, Jingren and Lin, Junyang},
	year = {2025},
}

@misc{yang_qwen3_2025,
	title = {Qwen3 {Technical} {Report}},
	url = {http://arxiv.org/abs/2505.09388},
	author = {Yang, An and Li, Anfeng and Yang, Baosong and Zhang, Beichen and Hui, Binyuan and Zheng, Bo and Yu, Bowen and Gao, Chang and Huang, Chengen and Lv, Chenxu and Zheng, Chujie and Liu, Dayiheng and Zhou, Fan and Huang, Fei and Hu, Feng and Ge, Hao and Wei, Haoran and Lin, Huan and Tang, Jialong and Yang, Jian and Tu, Jianhong and Zhang, Jianwei and Yang, Jianxin and Yang, Jiaxi and Zhou, Jing and Zhou, Jingren and Lin, Junyang and Dang, Kai and Bao, Keqin and Yang, Kexin and Yu, Le and Deng, Lianghao and Li, Mei and Xue, Mingfeng and Li, Mingze and Zhang, Pei and Wang, Peng and Zhu, Qin and Men, Rui and Gao, Ruize and Liu, Shixuan and Luo, Shuang and Li, Tianhao and Tang, Tianyi and Yin, Wenbiao and Ren, Xingzhang and Wang, Xinyu and Zhang, Xinyu and Ren, Xuancheng and Fan, Yang and Su, Yang and Zhang, Yichang and Zhang, Yinger and Wan, Yu and Liu, Yuqiong and Wang, Zekun and Cui, Zeyu and Zhang, Zhenru and Zhou, Zhipeng and Qiu, Zihan},
	year = {2025},
}

@misc{team_kimi_2025,
	title = {Kimi {K2}: {Open} {Agentic} {Intelligence}},
	url = {http://arxiv.org/abs/2507.20534},
	author = {Team, Kimi and Bai, Yifan and Bao, Yiping and Chen, Guanduo and Chen, Jiahao and Chen, Ningxin and Chen, Ruijue and Chen, Yanru and Chen, Yuankun and Chen, Yutian and Chen, Zhuofu and Cui, Jialei and Ding, Hao and Dong, Mengnan and Du, Angang and Du, Chenzhuang and Du, Dikang and Du, Yulun and Fan, Yu and Feng, Yichen and Fu, Kelin and Gao, Bofei and Gao, Hongcheng and Gao, Peizhong and Gao, Tong and Gu, Xinran and Guan, Longyu and Guo, Haiqing and Guo, Jianhang and Hu, Hao and Hao, Xiaoru and He, Tianhong and He, Weiran and He, Wenyang and Hong, Chao and Hu, Yangyang and Hu, Zhenxing and Huang, Weixiao and Huang, Zhiqi and Huang, Zihao and Jiang, Tao and Jiang, Zhejun and Jin, Xinyi and Kang, Yongsheng and Lai, Guokun and Li, Cheng and Li, Fang and Li, Haoyang and Li, Ming and Li, Wentao and Li, Yanhao and Li, Yiwei and Li, Zhaowei and Li, Zheming and Lin, Hongzhan and Lin, Xiaohan and Lin, Zongyu and Liu, Chengyin and Liu, Chenyu and Liu, Hongzhang and Liu, Jingyuan and Liu, Junqi and Liu, Liang and Liu, Shaowei and Liu, T. Y. and Liu, Tianwei and Liu, Weizhou and Liu, Yangyang and Liu, Yibo and Liu, Yiping and Liu, Yue and Liu, Zhengying and Lu, Enzhe and Lu, Lijun and Ma, Shengling and Ma, Xinyu and Ma, Yingwei and Mao, Shaoguang and Mei, Jie and Men, Xin and Miao, Yibo and Pan, Siyuan and Peng, Yebo and Qin, Ruoyu and Qu, Bowen and Shang, Zeyu and Shi, Lidong and Shi, Shengyuan and Song, Feifan and Su, Jianlin and Su, Zhengyuan and Sun, Xinjie and Sung, Flood and Tang, Heyi and Tao, Jiawen and Teng, Qifeng and Wang, Chensi and Wang, Dinglu and Wang, Feng and Wang, Haiming and Wang, Jianzhou and Wang, Jiaxing and Wang, Jinhong and Wang, Shengjie and Wang, Shuyi and Wang, Yao and Wang, Yejie and Wang, Yiqin and Wang, Yuxin and Wang, Yuzhi and Wang, Zhaoji and Wang, Zhengtao and Wang, Zhexu and Wei, Chu and Wei, Qianqian and Wu, Wenhao and Wu, Xingzhe and Wu, Yuxin and Xiao, Chenjun and Xie, Xiaotong and Xiong, Weimin and Xu, Boyu and Xu, Jing and Xu, Jinjing and Xu, L. H. and Xu, Lin and Xu, Suting and Xu, Weixin and Xu, Xinran and Xu, Yangchuan and Xu, Ziyao and Yan, Junjie and Yan, Yuzi and Yang, Xiaofei and Yang, Ying and Yang, Zhen and Yang, Zhilin and Yang, Zonghan and Yao, Haotian and Yao, Xingcheng and Ye, Wenjie and Ye, Zhuorui and Yin, Bohong and Yu, Longhui and Yuan, Enming and Yuan, Hongbang and Yuan, Mengjie and Zhan, Haobing and Zhang, Dehao and Zhang, Hao and Zhang, Wanlu and Zhang, Xiaobin and Zhang, Yangkun and Zhang, Yizhi and Zhang, Yongting and Zhang, Yu and Zhang, Yutao and Zhang, Yutong and Zhang, Zheng and Zhao, Haotian and Zhao, Yikai and Zheng, Huabin and Zheng, Shaojie and Zhou, Jianren and Zhou, Xinyu and Zhou, Zaida and Zhu, Zhen and Zhuang, Weiyu and Zu, Xinxing},
	year = {2025},
}

@misc{deepseek-ai_deepseek-v3_2025,
	title = {{DeepSeek}-{V3} {Technical} {Report}},
	url = {http://arxiv.org/abs/2412.19437},
	author = {DeepSeek-AI and Liu, Aixin and Feng, Bei and Xue, Bing and Wang, Bingxuan and Wu, Bochao and Lu, Chengda and Zhao, Chenggang and Deng, Chengqi and Zhang, Chenyu and Ruan, Chong and Dai, Damai and Guo, Daya and Yang, Dejian and Chen, Deli and Ji, Dongjie and Li, Erhang and Lin, Fangyun and Dai, Fucong and Luo, Fuli and Hao, Guangbo and Chen, Guanting and Li, Guowei and Zhang, H. and Bao, Han and Xu, Hanwei and Wang, Haocheng and Zhang, Haowei and Ding, Honghui and Xin, Huajian and Gao, Huazuo and Li, Hui and Qu, Hui and Cai, J. L. and Liang, Jian and Guo, Jianzhong and Ni, Jiaqi and Li, Jiashi and Wang, Jiawei and Chen, Jin and Chen, Jingchang and Yuan, Jingyang and Qiu, Junjie and Li, Junlong and Song, Junxiao and Dong, Kai and Hu, Kai and Gao, Kaige and Guan, Kang and Huang, Kexin and Yu, Kuai and Wang, Lean and Zhang, Lecong and Xu, Lei and Xia, Leyi and Zhao, Liang and Wang, Litong and Zhang, Liyue and Li, Meng and Wang, Miaojun and Zhang, Mingchuan and Zhang, Minghua and Tang, Minghui and Li, Mingming and Tian, Ning and Huang, Panpan and Wang, Peiyi and Zhang, Peng and Wang, Qiancheng and Zhu, Qihao and Chen, Qinyu and Du, Qiushi and Chen, R. J. and Jin, R. L. and Ge, Ruiqi and Zhang, Ruisong and Pan, Ruizhe and Wang, Runji and Xu, Runxin and Zhang, Ruoyu and Chen, Ruyi and Li, S. S. and Lu, Shanghao and Zhou, Shangyan and Chen, Shanhuang and Wu, Shaoqing and Ye, Shengfeng and Ye, Shengfeng and Ma, Shirong and Wang, Shiyu and Zhou, Shuang and Yu, Shuiping and Zhou, Shunfeng and Pan, Shuting and Wang, T. and Yun, Tao and Pei, Tian and Sun, Tianyu and Xiao, W. L. and Zeng, Wangding and Zhao, Wanjia and An, Wei and Liu, Wen and Liang, Wenfeng and Gao, Wenjun and Yu, Wenqin and Zhang, Wentao and Li, X. Q. and Jin, Xiangyue and Wang, Xianzu and Bi, Xiao and Liu, Xiaodong and Wang, Xiaohan and Shen, Xiaojin and Chen, Xiaokang and Zhang, Xiaokang and Chen, Xiaosha and Nie, Xiaotao and Sun, Xiaowen and Wang, Xiaoxiang and Cheng, Xin and Liu, Xin and Xie, Xin and Liu, Xingchao and Yu, Xingkai and Song, Xinnan and Shan, Xinxia and Zhou, Xinyi and Yang, Xinyu and Li, Xinyuan and Su, Xuecheng and Lin, Xuheng and Li, Y. K. and Wang, Y. Q. and Wei, Y. X. and Zhu, Y. X. and Zhang, Yang and Xu, Yanhong and Xu, Yanhong and Huang, Yanping and Li, Yao and Zhao, Yao and Sun, Yaofeng and Li, Yaohui and Wang, Yaohui and Yu, Yi and Zheng, Yi and Zhang, Yichao and Shi, Yifan and Xiong, Yiliang and He, Ying and Tang, Ying and Piao, Yishi and Wang, Yisong and Tan, Yixuan and Ma, Yiyang and Liu, Yiyuan and Guo, Yongqiang and Wu, Yu and Ou, Yuan and Zhu, Yuchen and Wang, Yuduan and Gong, Yue and Zou, Yuheng and He, Yujia and Zha, Yukun and Xiong, Yunfan and Ma, Yunxian and Yan, Yuting and Luo, Yuxiang and You, Yuxiang and Liu, Yuxuan and Zhou, Yuyang and Wu, Z. F. and Ren, Z. Z. and Ren, Zehui and Sha, Zhangli and Fu, Zhe and Xu, Zhean and Huang, Zhen and Zhang, Zhen and Xie, Zhenda and Zhang, Zhengyan and Hao, Zhewen and Gou, Zhibin and Ma, Zhicheng and Yan, Zhigang and Shao, Zhihong and Xu, Zhipeng and Wu, Zhiyu and Zhang, Zhongyu and Li, Zhuoshu and Gu, Zihui and Zhu, Zijia and Liu, Zijun and Li, Zilin and Xie, Ziwei and Song, Ziyang and Gao, Ziyi and Pan, Zizheng},
	year = {2025},
}

@inproceedings{berges_memory_2025,
	title = {Memory {Layers} at {Scale}},
	url = {https://proceedings.mlr.press/v267/berges25a.html},
	booktitle = {Proceedings of the 42nd {International} {Conference} on {Machine} {Learning}},
	author = {Berges, Vincent-Pierre and Oguz, Barlas and Haziza, Daniel and Yih, Wen-Tau and Zettlemoyer, Luke and Ghosh, Gargi},
	year = {2025},
	note = {ISSN: 2640-3498},
	pages = {3831--3842},
}

@inproceedings{huang_over-tokenized_2025,
	title = {Over-{Tokenized} {Transformer}: {Vocabulary} is {Generally} {Worth} {Scaling}},
	url = {https://proceedings.mlr.press/v267/huang25bb.html},
	booktitle = {Proceedings of the 42nd {International} {Conference} on {Machine} {Learning}},
	author = {Huang, Hongzhi and Zhu, Defa and Wu, Banggu and Zeng, Yutao and Wang, Ya and Min, Qiyang and Xun, Zhou},
	year = {2025},
	note = {ISSN: 2640-3498},
	pages = {26261--26282},
}

@inproceedings{shazeer_outrageously_2017,
	title = {Outrageously {Large} {Neural} {Networks}: {The} {Sparsely}-{Gated} {Mixture}-of-{Experts} {Layer}},
	url = {https://openreview.net/forum?id=B1ckMDqlg},
	booktitle = {{International} {Conference} on {Learning} {Representations} 2017},
	author = {Shazeer, Noam and Mirhoseini, Azalia and Maziarz, Krzysztof and Davis, Andy and Le, Quoc V. and Hinton, Geoffrey E. and Dean, Jeff},
	year = {2017},
}

@inproceedings{dai_deepseekmoe_2024,
	title = {{DeepSeekMoE}: {Towards} {Ultimate} {Expert} {Specialization} in {Mixture}-of-{Experts} {Language} {Models}},
	url = {https://aclanthology.org/2024.acl-long.70/},
	booktitle = {Proceedings of the 62nd {Annual} {Meeting} of the {Association} for {Computational} {Linguistics} ({Volume} 1: {Long} {Papers})},
	author = {Dai, Damai and Deng, Chengqi and Zhao, Chenggang and Xu, R.x. and Gao, Huazuo and Chen, Deli and Li, Jiashi and Zeng, Wangding and Yu, Xingkai and Wu, Y. and Xie, Zhenda and Li, Y.k. and Huang, Panpan and Luo, Fuli and Ruan, Chong and Sui, Zhifang and Liang, Wenfeng},
	year = {2024},
	pages = {1280--1297},
}

@misc{yu_scaling_2025,
	title = {Scaling {Embedding} {Layers} in {Language} {Models}},
	url = {http://arxiv.org/abs/2502.01637},
	author = {Yu, Da and Cohen, Edith and Ghazi, Badih and Huang, Yangsibo and Kamath, Pritish and Kumar, Ravi and Liu, Daogao and Zhang, Chiyuan},
	year = {2025},
}

@article{huang_ultra-sparse_2025,
	title = {Ultra-{Sparse} {Memory} {Network}},
	volume = {2025},
	url = {https://proceedings.iclr.cc/paper_files/paper/2025/hash/d78d68cae595fabadd187b583ee8708e-Abstract-Conference.html},
	journal = {International Conference on Representation Learning},
	author = {Huang, Zihao and Min, Qiyang and Huang, Hongzhi and Zeng, Yutao and Zhu, Defa and Guo, Ran and Xun, Zhou},
	year = {2025},
	pages = {86557--86575},
}

@misc{huang_ultramemv2_2025,
	title = {{UltraMemV2}: {Memory} {Networks} {Scaling} to {120B} {Parameters} with {Superior} {Long}-{Context} {Learning}},
	url = {http://arxiv.org/abs/2508.18756},
	author = {Huang, Zihao and Bao, Yu and Min, Qiyang and Chen, Siyan and Guo, Ran and Huang, Hongzhi and Zhu, Defa and Zeng, Yutao and Wu, Banggu and Zhou, Xun and Qiao, Siyuan},
	month = aug,
	year = {2025},
}

@inproceedings{lample_large_2019,
	title = {Large {Memory} {Layers} with {Product} {Keys}},
	volume = {32},
	url = {https://proceedings.neurips.cc/paper/2019/hash/9d8df73a3cfbf3c5b47bc9b50f214aff-Abstract.html},
	booktitle = {Advances in {Neural} {Information} {Processing} {Systems}},
	author = {Lample, Guillaume and Sablayrolles, Alexandre and Ranzato, Marc' Aurelio and Denoyer, Ludovic and Jegou, Herve},
	year = {2019},
}

@misc{xue_powerinfer-2_2024,
	title = {{PowerInfer}-2: {Fast} {Large} {Language} {Model} {Inference} on a {Smartphone}},
	url = {http://arxiv.org/abs/2406.06282},
	author = {Xue, Zhenliang and Song, Yixin and Mi, Zeyu and Zheng, Xinrui and Xia, Yubin and Chen, Haibo},
	year = {2024},
}

@inproceedings{qiu_demons_2025,
	title = {Demons in the {Detail}: {On} {Implementing} {Load} {Balancing} {Loss} for {Training} {Specialized} {Mixture}-of-{Expert} {Models}},
	isbn = {979-8-89176-251-0},
	url = {https://aclanthology.org/2025.acl-long.249/},
	booktitle = {Proceedings of the 63rd {Annual} {Meeting} of the {Association} for {Computational} {Linguistics} ({Volume} 1: {Long} {Papers})},
	author = {Qiu, Zihan and Huang, Zeyu and Zheng, Bo and Wen, Kaiyue and Wang, Zekun and Men, Rui and Titov, Ivan and Liu, Dayiheng and Zhou, Jingren and Lin, Junyang},
	year = {2025},
	pages = {5005--5018},
}

@inproceedings{ludziejewski_scaling_2024,
	title = {Scaling {Laws} for {Fine}-{Grained} {Mixture} of {Experts}},
	url = {https://proceedings.mlr.press/v235/ludziejewski24a.html},
	issn = {2640-3498},
	booktitle = {Proceedings of the 41st {International} {Conference} on {Machine} {Learning}},
	author = {Ludziejewski, Jan and Krajewski, Jakub and Adamczewski, Kamil and Pióro, Maciej and Krutul, Michał and Antoniak, Szymon and Ciebiera, Kamil and Król, Krystian and Odrzygóźdź, Tomasz and Sankowski, Piotr and Cygan, Marek and Jaszczur, Sebastian},
	year = {2024},
	pages = {33270--33288},
}

@misc{shazeer_glu_2020,
	title = {{GLU} {Variants} {Improve} {Transformer}},
	url = {http://arxiv.org/abs/2002.05202},
	author = {Shazeer, Noam},
	year = {2020},
}

@inproceedings{xiong_layer_2020,
	title = {On {Layer} {Normalization} in the {Transformer} {Architecture}},
	url = {https://proceedings.mlr.press/v119/xiong20b.html},
	issn = {2640-3498},
	booktitle = {Proceedings of the 37th {International} {Conference} on {Machine} {Learning}},
	author = {Xiong, Ruibin and Yang, Yunchang and He, Di and Zheng, Kai and Zheng, Shuxin and Xing, Chen and Zhang, Huishuai and Lan, Yanyan and Wang, Liwei and Liu, Tieyan},
	year = {2020},
	pages = {10524--10533},
}

@inproceedings{zhang_root_2019,
	title = {Root {Mean} {Square} {Layer} {Normalization}},
	volume = {32},
	url = {https://proceedings.neurips.cc/paper/2019/hash/1e8a19426224ca89e83cef47f1e7f53b-Abstract.html},
	booktitle = {Advances in {Neural} {Information} {Processing} {Systems}},
	author = {Zhang, Biao and Sennrich, Rico},
	year = {2019},
}

@inproceedings{loshchilov_adamw_2019,
  author       = {Ilya Loshchilov and Frank Hutter},
  title        = {Decoupled Weight Decay Regularization},
  booktitle    = {7th International Conference on Learning Representations, {ICLR} 2019,},
  year         = {2019},
  url          = {https://openreview.net/forum?id=Bkg6RiCqY7},
}

\appendix

\newpage
\section{Pseudocode}

The pseudocode for MoKLV in training mode is given below. The definitions of some modules that are not the focus of this article, such as Attention, RotaryEmb, and Config, are omitted here.
\renewcommand{\theFancyVerbLine}{\ttfamily\textcolor[rgb]{0.5,0.5,0.5}{\scriptsize\arabic{FancyVerbLine}}}
\begin{minted}[
    linenos,
    numbersep=5pt,
    frame=lines,
    fontsize=\fontsize{10pt}{12pt}\selectfont,
    framesep=4mm,
]{python}
import torch
from torch import nn
from torch.nn import functional as F


class MLP(nn.Module):
    def __init__(self, config: Config, key_expert: bool = False):
        super().__init__()
        h_size = config.hidden_size
        i_size = config.intermediate_size
        o_size = config.key_size if key_expert else h_size

        self.ffn_gate = nn.Linear(h_size, i_size, bias=False)
        self.ffn_up = nn.Linear(h_size, i_size, bias=False)
        self.ffn_down = nn.Linear(i_size, o_size, bias=False)

    def forward(self, x: torch.Tensor):
        x1 = self.ffn_gate(x)
        x2 = self.ffn_up(x)
        x3 = F.silu(x1) * x2
        y = self.ffn_down(x3)
        return y


class Layer(nn.Module):
    def __init__(self, config: Config, has_experts: bool):
        super().__init__()
        d = config.hidden_size
        dk = config.key_size
        n = config.num_experts

        self.rotary_emb = RotaryEmb(config)
        self.attention = Attention(config, self.rotary_emb)
        self.ffn = MLP(config)
        self.attn_layer_norm = nn.RMSNorm(d)
        self.ffn_layer_norm = nn.RMSNorm(d)
        self.has_experts = has_experts
        self.qk_scale_factor = config.qk_scale_factor
        self.expert_topk = config.expert_topk

        if has_experts:
            self.query_proj = nn.Linear(d, dk, bias=False)
            self.key_experts = nn.ModuleList([MLP(config, True) for _ in range(n)])
            self.value_experts = nn.ModuleList([MLP(config) for _ in range(n)])
            self.vocab_emb_layer_norm = nn.RMSNorm(d)
            self.expert_key_norm = nn.RMSNorm(dk)
            self.expert_value_norm = nn.RMSNorm(d)
            self.router = nn.Parameter(torch.randn(d, n))
            self.new_router = nn.Parameter(torch.randn(d, n))
            self.gate = nn.Parameter(torch.randn(d, 1))
            self.new_gate = nn.Parameter(torch.randn(d, 1))

    def forward(
        self,
        x_input: torch.Tensor,
        vocab_emb: torch.Tensor,
        sliding_window_mask: torch.Tensor,
    ):
        x = self.attn_layer_norm(x_input)
        x_ffn_input = self.attention(x) + x_input
        x_ffn = self.ffn_layer_norm(x_ffn_input)
        y = self.ffn(x_ffn) + x_ffn_input

        if self.has_experts:
            expert_query = self.query_proj(x_ffn).unsqueeze(2)  # (b, s, 1, dk)
            vocab_emb = self.vocab_emb_layer_norm(vocab_emb)
            expert_key = torch.stack(
                [f(vocab_emb) for f in self.key_experts], 2
            )  # (b, s, n, dk)
            expert_key = self.expert_key_norm(expert_key)
            expert_value = torch.stack(
                [f(vocab_emb) for f in self.value_experts], 2
            )  # (b, s, n, d)

            """The original expert output in MoLE"""
            router_score = x_ffn @ self.router  # (b, s, n)
            qk_score = (
                expert_query @ expert_key.transpose(-2, -1) * self.qk_scale_factor
            )  # (b, s, 1, dk) @ (b, s, dk, n) -> (b, s, 1, n)
            score = router_score + qk_score.squeeze(2)  # (b, s, n)
            score = score.softmax(-1).unsqueeze(-1)  # (b, s, n, 1)
            gate = torch.sigmoid(x_ffn @ self.gate)  # (b, s, 1)
            expert_output = (expert_value * score).sum(2) * gate  # (b, s, d)

            """The New expert output added in MoLKV"""
            new_router_score = x_ffn @ self.new_router  # (b, s, n)
            expert_query_r = self.rotary_emb(expert_query)
            expert_key_r = self.rotary_emb(expert_key)
            new_qk_score = (
                expert_query_r.transpose(1, 2)
                @ expert_key_r.permute(0, 2, 3, 1)
                * self.qk_scale_factor
            )  # (b, 1, s, dk) @ (b, n, dk, s) -> (b, n, s, s)
            new_score = (
                new_router_score.transpose(1, 2).unsqueeze(-1) + new_qk_score
            )  # (b, n, s, s)

            new_score = new_score.masked_fill(~sliding_window_mask, float("-inf"))
            new_score = new_score.permute(0, 2, 3, 1).flatten(2)  # (b, s, s * n)
            indices = new_score.topk(self.expert_topk, -1, sorted=False).indices
            topk_mask = torch.zeros_like(new_score, dtype=bool)
            topk_mask.scatter_(-1, indices, True)
            new_score = new_score.masked_fill(~topk_mask, float("-inf"))
            new_score = new_score.softmax(-1)
            new_gate = torch.sigmoid(x_ffn @ self.new_gate)  # (b, s, 1)
            new_expert_output = (
                new_score
                @ self.expert_value_norm(expert_value).flatten(1, 2)
                * new_gate
            )  # (b, s, s * n) @ (b, s * n, d) -> (b, s, d)

            y = y + expert_output + new_expert_output

        return y
\end{minted}

\section{Training Hyperparameters}
\begin{table}[h]
\small
\centering
\begin{tabular}{lr}
    \toprule
    Configuration Key & Value \\
    \midrule
    seq\_length & 2048 \\
    batch\_size & 8 \\
    gradient\_accumulatio\_steps & 30 \\
    effective\_batch\_size & 240 \\
    training\_steps & 20000 \\
    warmup\_steps & 200 \\
    learning\_rate & 3e-4 \\
    lr\_decay\_scheduler & cosine \\
    min\_lr & 3e-6 \\
    weight\_decay & 0.1 \\
    adam\_betas & [0.9, 0.95] \\
    grad\_clip & 1.0 \\
    layer\_norm\_type & RMSNorm \\
    layer\_norm\_eps & 1e-8 \\
    optimizer\_type & AdamW \\
    optimizer\_eps & 1e-8 \\
    mixed\_precision & True \\
    pos\_emb & RoPE \\
    pos\_emb\_theta & 10000 \\
    weight\_tying & False \\
    init\_dist & trunc normal \\
    init\_std & 0.02 \\
\bottomrule
\end{tabular}
\label{tab:training config}
\end{table}

\end{document}